\documentclass[conference]{IEEEtran}
\IEEEoverridecommandlockouts
\usepackage{cite}
\usepackage{amsmath,amssymb,amsfonts}
\usepackage{algorithmic}
\usepackage{graphicx}
\usepackage{textcomp}
\usepackage{xcolor}

\usepackage{hyperref}
\usepackage{xfrac}
\usepackage{siunitx}

\def\BibTeX{{\rm B\kern-.05em{\sc i\kern-.025em b}\kern-.08em
    T\kern-.1667em\lower.7ex\hbox{E}\kern-.125emX}}
\begin{document}

\title{Deep Learning Methods for Detecting Thermal Runaway Events in Battery Production Lines}

\author{\IEEEauthorblockN{Athanasios Athanasopoulos, Mat\'u\v{s} Mihal\'ak, Marcin Pietrasik*\thanks{*Corresponding author: \href{mailto:marcin.pietrasik@maastrichtuniversity.nl}{marcin.pietrasik@maastrichtuniversity.nl}} }
\IEEEauthorblockA{{Department of Advanced Computing Sciences} \\
Maastricht University \\
Maastricht, The Netherlands}
}

© 2025 IEEE. Personal use of this material is permitted. Permission from IEEE must be
obtained for all other uses, in any current or future media, including
reprinting/republishing this material for advertising or promotional purposes, creating new
collective works, for resale or redistribution to servers or lists, or reuse of any copyrighted
component of this work in other works.

This paper has been published in the \textit{8th IEEE Conference on Industrial Cyber-Physical Systems (ICPS)} in Emden, Germany, May 12-15, 2025.

A. Athanasopoulos, M. Mihalák and M. Pietrasik, ``Deep Learning Methods for Detecting Thermal Runaway Events in Battery Production Lines,'' \textit{2025 IEEE 8th International Conference on Industrial Cyber-Physical Systems (ICPS)}, Emden, Germany, 2025, pp. 1-6, doi: 10.1109/ICPS65515.2025.11087884.

\clearpage

\maketitle

\begin{abstract}

One of the key safety considerations of battery manufacturing is thermal runaway, the uncontrolled increase in temperature which can lead to fires, explosions, and emissions of toxic gasses. As such, development of automated systems capable of detecting such events is of considerable importance in both academic and industrial contexts. In this work, we investigate the use of deep learning for detecting thermal runaway in the battery production line of VDL Nedcar, a Dutch automobile manufacturer. Specifically, we collect data from the production line to represent both baseline (non thermal runaway) and thermal runaway conditions. Thermal runaway was simulated through the use of external heat and smoke sources. The data consisted of both optical and thermal images which were then preprocessed and fused before serving as input to our models. In this regard, we evaluated three deep-learning models widely used in computer vision including shallow convolutional neural networks, residual neural networks, and vision transformers on two performance metrics. Furthermore, we evaluated these models using explainability methods to gain insight into their ability to capture the relevant feature information from their inputs. The obtained results indicate that the use of deep learning is a viable approach to thermal runaway detection in battery production lines.

\end{abstract}

\begin{IEEEkeywords}
thermal runaway, deep learning, data fusion, classification, manufacturing
\end{IEEEkeywords}

\section{Introduction}

Thermal runaway is the process in which an increase in temperature triggers a self-reinforcing cycle of exothermic reactions, leading to a rapid and uncontrollable rise in temperature \cite{meta_learning_thermal_runaway_forecasting,data_driven_anomaly_detection_batteries_unsupervised_clustering,ors2025machine}. In the context of battery manufacturing, this rise in temperature occurs within the battery cell on a production line, potentially leading to fires, explosions, or system failures if not promptly detected and mitigated. As such, thermal runaway presents a critical safety consideration and challenge for manufacturers and its early detection is of critical importance to minimize damages \cite{review_thermal_runaway_prevention_mitigation_strategies_lithium_ion_batteries}.
Common detection systems rely on temperature thresholds and gas detection \cite{kong2023review}. These approaches are not sufficient, however, as it has been observed that in thermal runaway events, smoke oftentimes precedes heat rise \cite{chai2024experimental} while production realities prevent smoke sensors from being proximally located to the emission source, thus delaying detection.

In this paper, we propose the use of deep learning methods with the data fusion of static optical and infrared battery images obtained during the production process, thus integrating physical processes with cyber systems. 
This allows us to leverage both smoke and heat information and potentially allow for earlier and more accurate detection. 
Specifically, we investigate the use of three common computer vision models on data designed to simulate thermal runaway events in a production environment.
To the best of our knowledge, this work is the first to leverage such methods for thermal runaway detection in the context of battery production lines.
We performed our work in collaboration with the automotive manufacturing company VDL Nedcar which provided us with the use case and access to their production line. 
The production line is modular and capable of building a variety of battery packs with minimal re-engineering effort. It is comprised of two automated stations, four manual stations, a rework station, and a loading station. 
Our work focused on detecting thermal runaway in the first automated station, which follows a manual pack preparation station. In this sense, it also acts as a check on the human labour performed earlier.
We collected data from the battery line for model evaluation and found that all three models tested were capable of achieving excellent results as per our performance metrics.
We thus conclude that deep learning methods are a viable approach for detecting thermal runaway events in the context of battery production lines.

Our paper continues with a brief overview of the related work in Section \ref{sec:related_works} before outlining the materials and methods used in Section \ref{sec:materials_and_methods}. The results of these methods are captured and discussed in Section \ref{sec:evaluation}. Section \ref{sec:conclusion} concludes the paper and proposes avenues for future research.

\section{Related Work}\label{sec:related_works}

Recent advances in deep learning have spurred increasing academic and industry interest in its application for the purpose of thermal runaway detection.
Perhaps the work which bears the most resemblance to our own is that of Ding et al. \cite{meta_learning_thermal_runaway_forecasting}. Similar to our work, their method uses a data fusion approach in which thermal images are used with voltage and temperature data as input to a neural network. The approach achieved encouraging results on simulated samples and the authors noted that incorporating infrared images increased model performance.
In a related work, Hong et al. \cite{Fault_prognosis_voltage_batteries_lstm} did not use optical image data but rather a time series of voltage measured in the battery cells. This data was then used to train a long short-term memory neural network to forecast when a fault will occur in a battery cell system.
In another time series forecasting approach, Li et al. \cite{data_driven_anomaly_detection_batteries_unsupervised_clustering} approached the problem using unsupervised learning. Specifically, they employed k-Shape clustering on thermal data from sensors near the battery to construct clusters and conclude that this approach is more robust to data loss compared to using optical image data.
More recently, Örs and Javani \cite{ors2025machine} applied various common machine learning algorithms to predict thermal runaway using external heat and force information to obtain reliable results.
In another approach that leverages external information, Tam et al. \cite{tam2024development} developed a detection model for early-stage thermal runaway using acoustic data.
These works differ from our own in that they detect thermal runaway in already assembled and operating batteries, not during their manufacturing process.
Furthermore, none of the aforementioned works have integrated explainability into their approaches. This presents a challenge as explainability is a key factor in the deployment of deep learning methods in industrial settings \cite{agostinho2023explainability}.
With this in mind, there has been work done in integrating importance heatmaps in the tangential fields of fire \cite{image_fire_detection_algorithms_cnn} and smoke \cite{weighted_ensemble_approach_smoke_like_classification} detection.
Comprehensive reviews of the state-of-the-art machine learning and deep learning approaches for thermal management in batteries is provided by Li et al. \cite{li2023machine} and Qi et al. \cite{qi2024advanced}, respectively.

\section{Materials and Methods} 
\label{sec:materials_and_methods}

\begin{figure*}[t!] 
    \centering
    \includegraphics[width=0.22\textwidth]{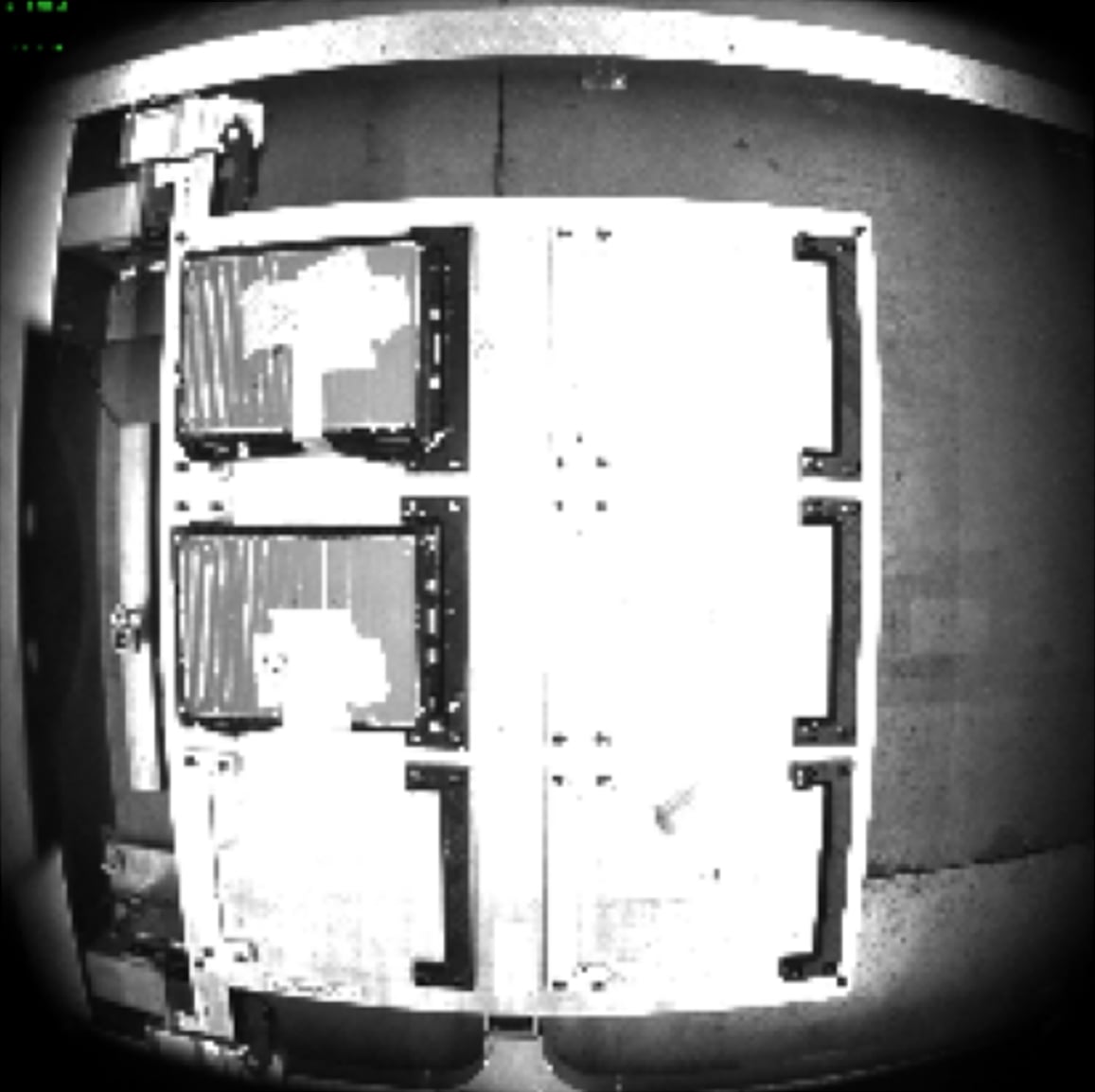}
    \hspace{0.2cm}
    \includegraphics[width=0.22\textwidth]{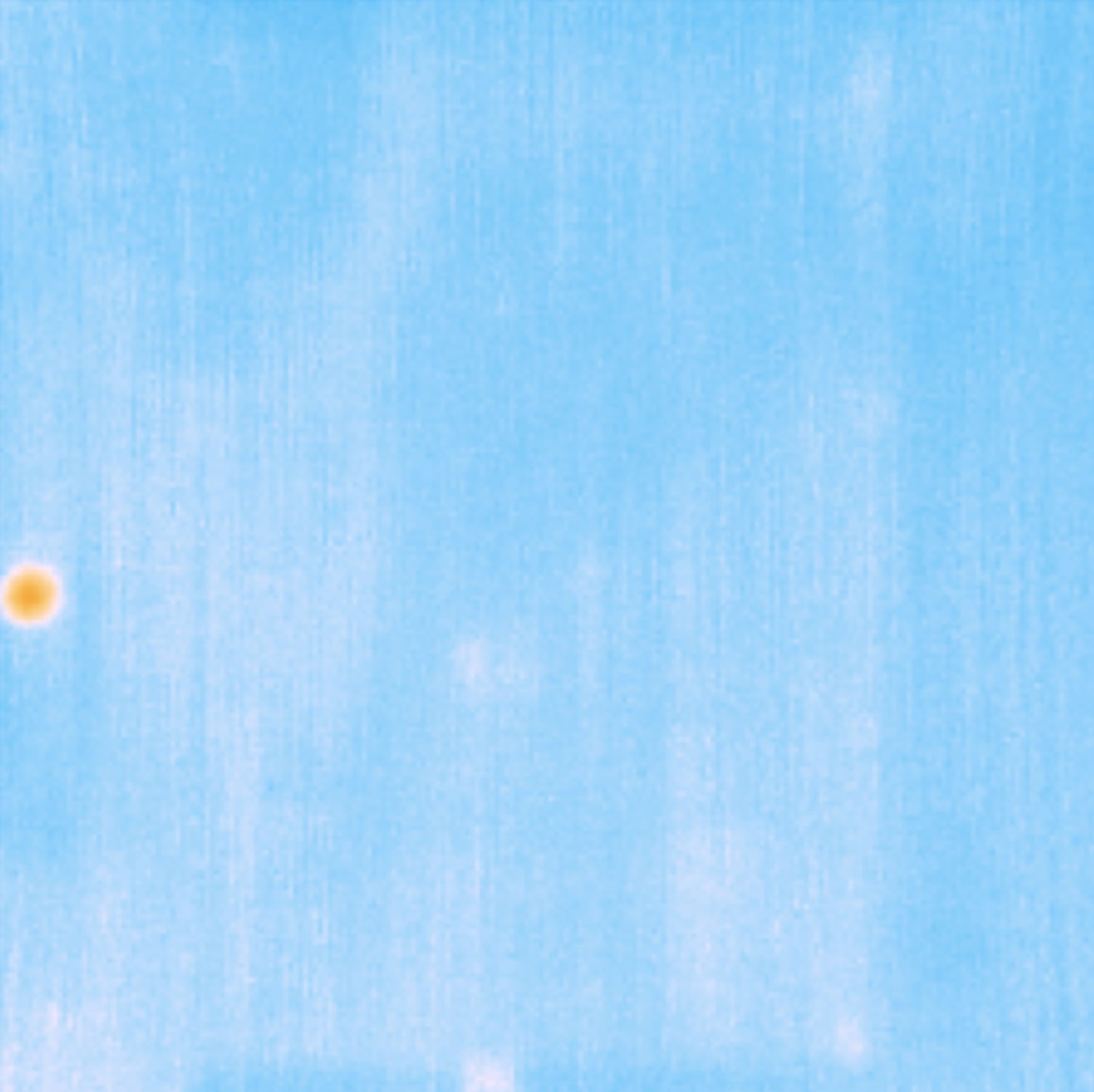}
    \hspace{1cm}
    \includegraphics[width=0.22\textwidth]{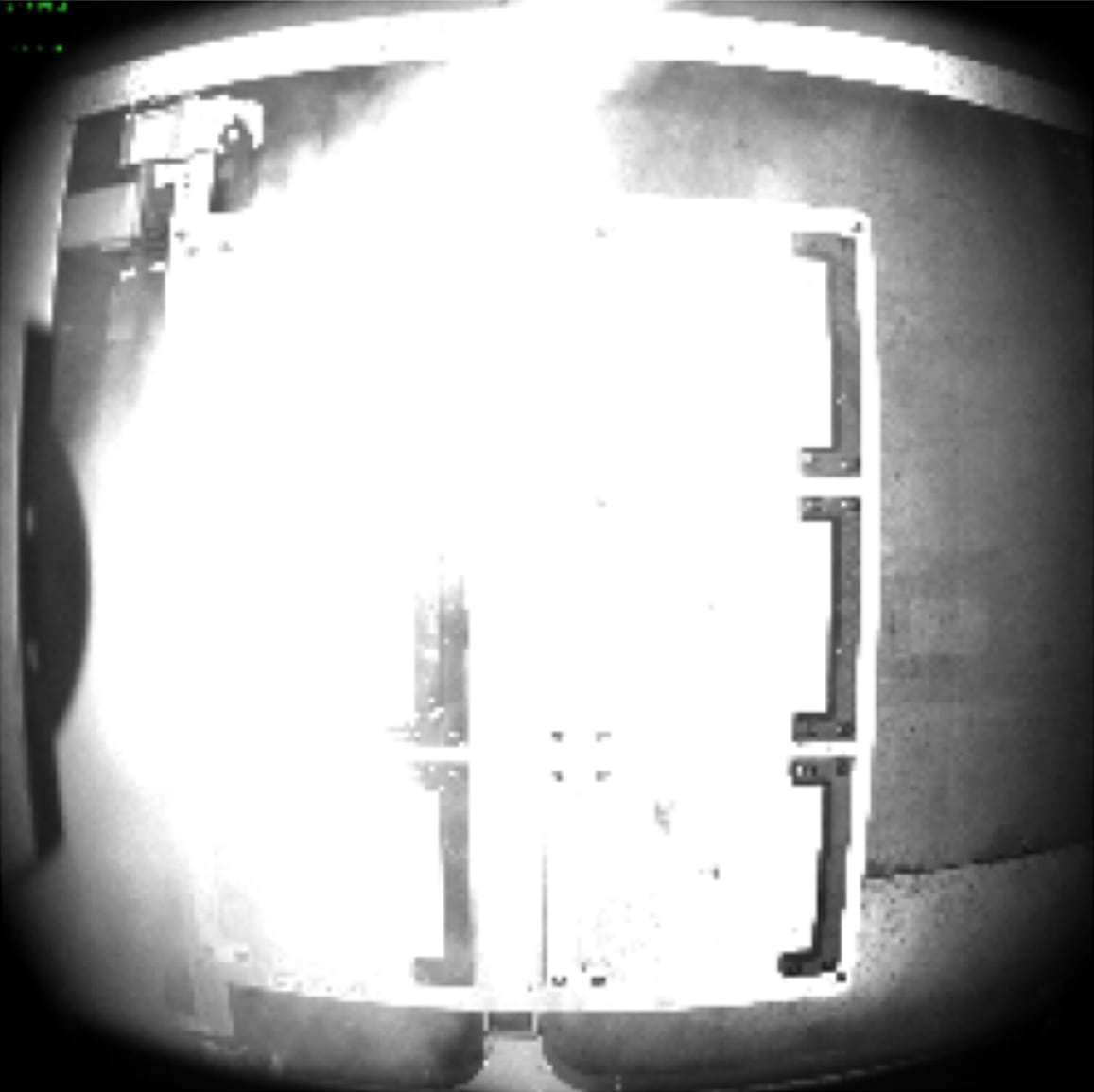}
    \hspace{0.2cm}
    \includegraphics[width=0.22\textwidth]{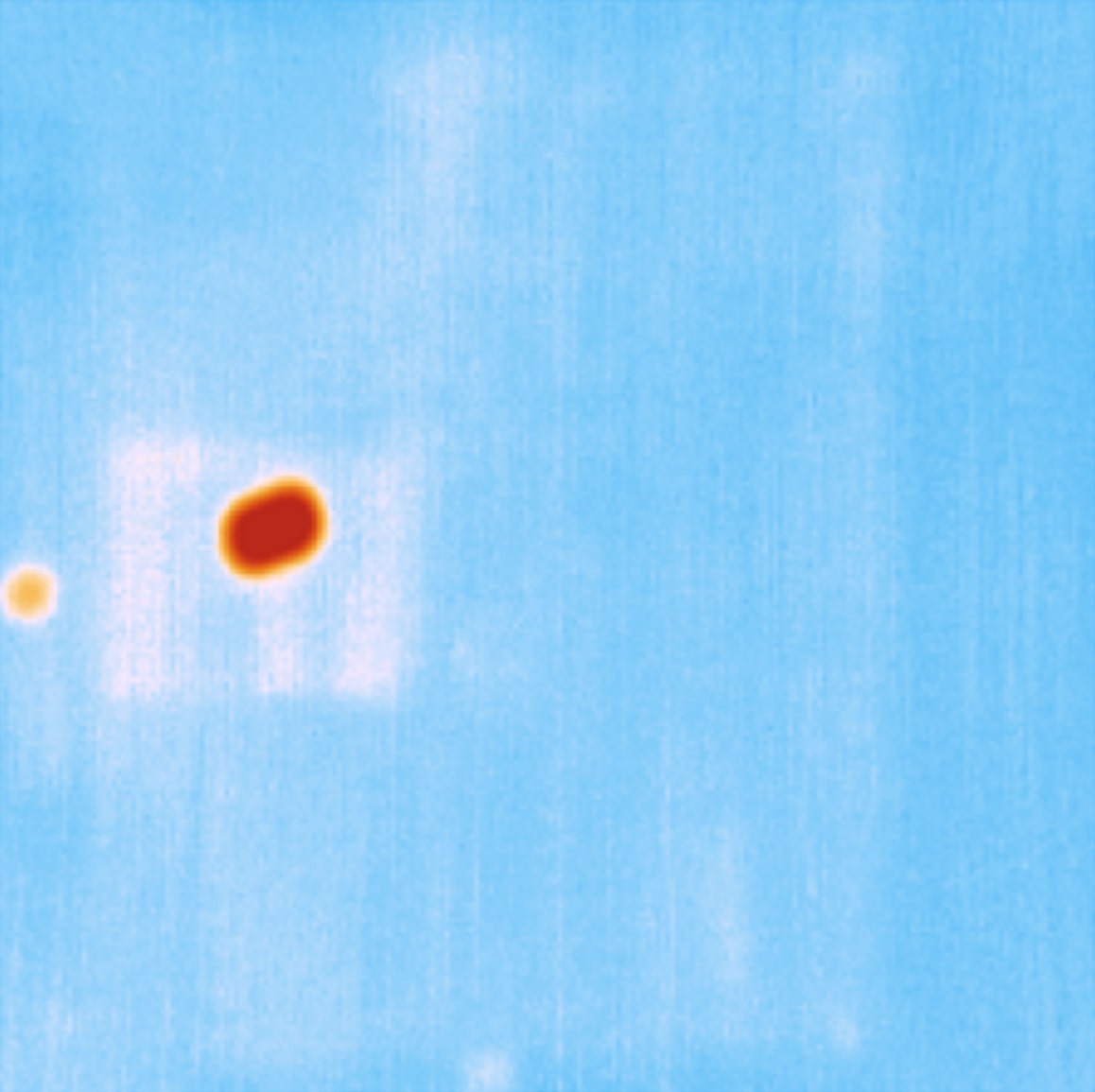}\\
    \vspace{0.1cm}
    (a) \hspace{9.1cm} (b) \\
    \vspace{0.2cm}
    \includegraphics[width=0.22\textwidth]{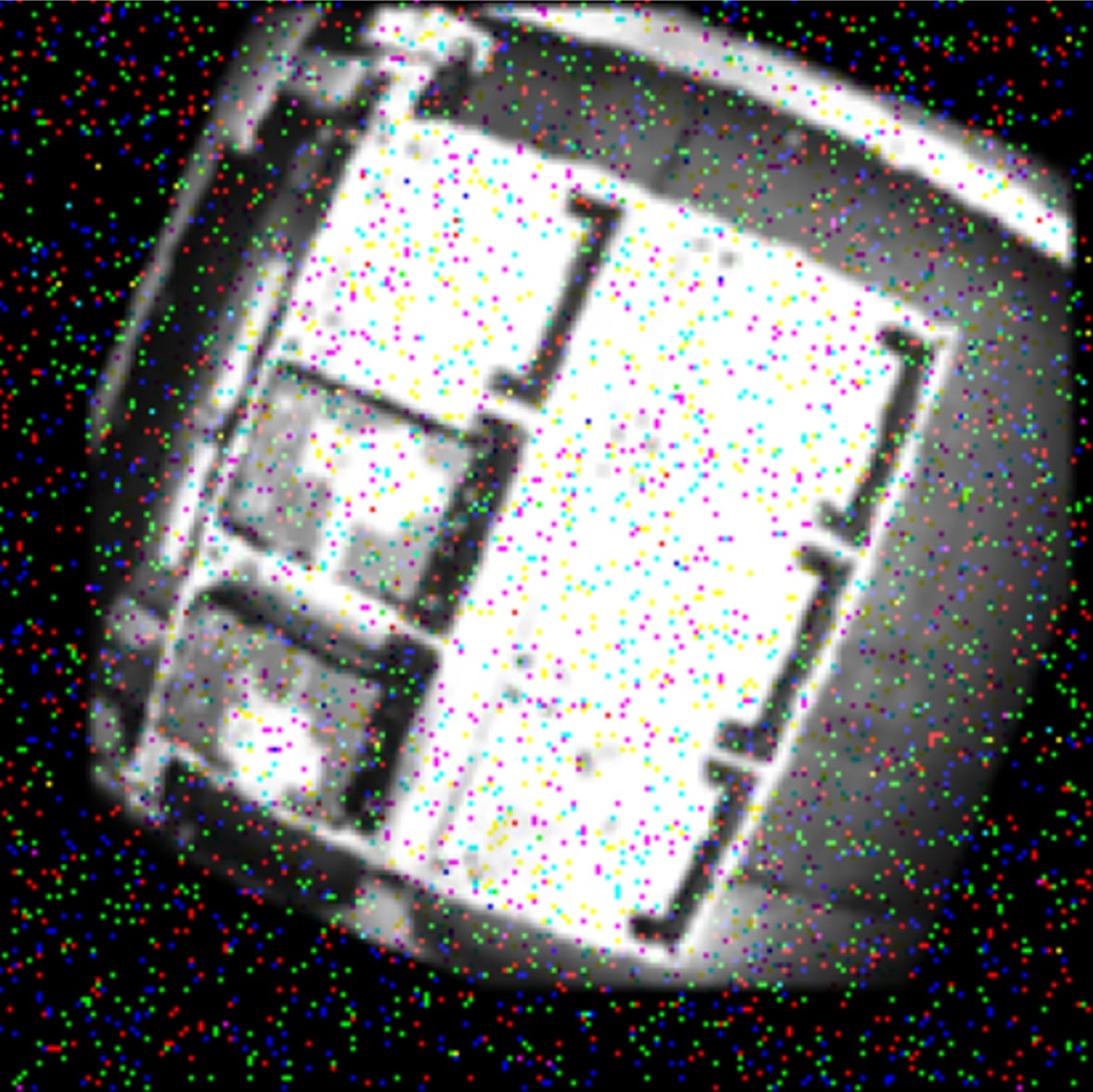}
    \hspace{0.2cm}
    \includegraphics[width=0.22\textwidth]{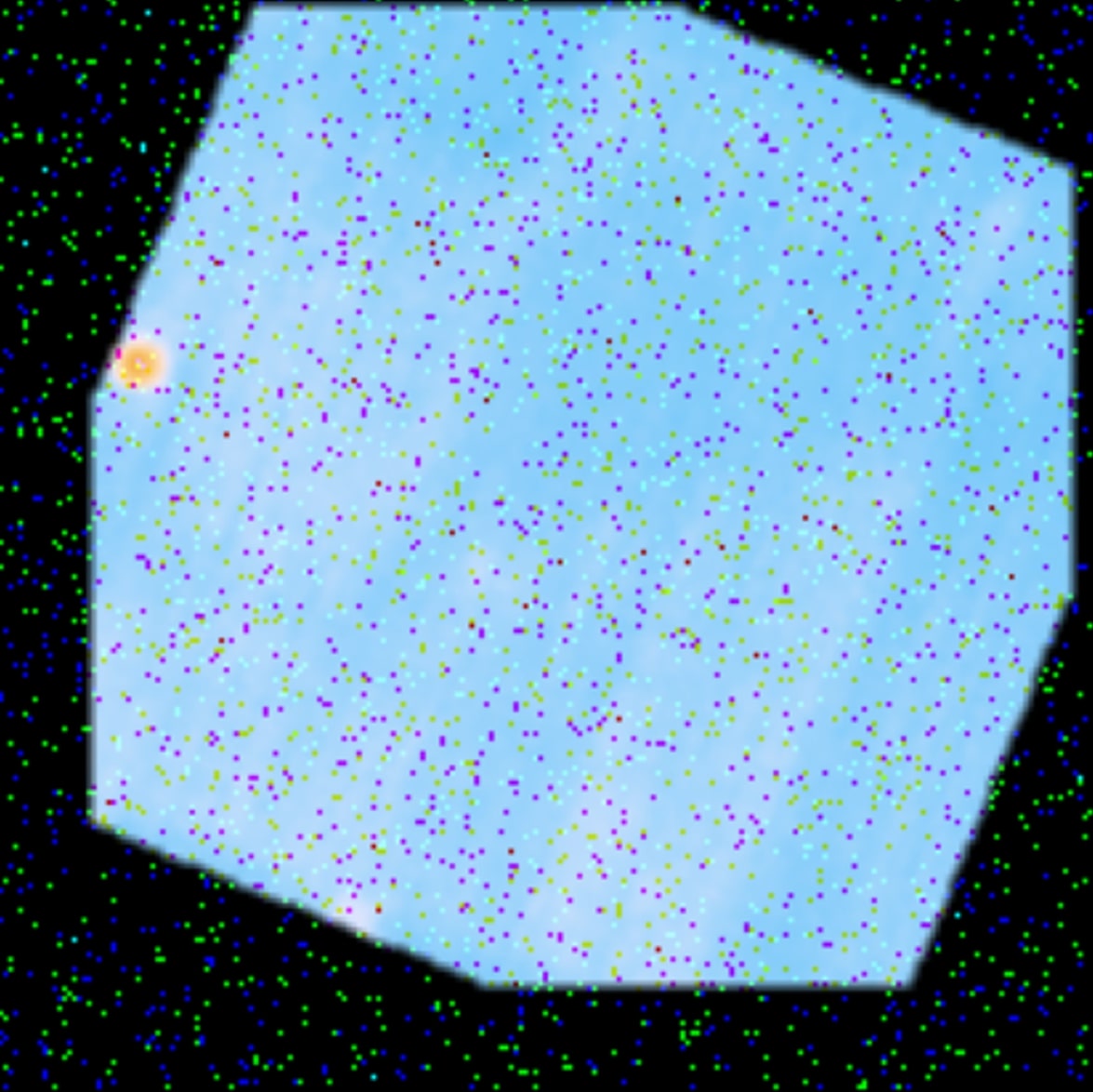}
    \hspace{1cm}
    \includegraphics[width=0.22\textwidth]{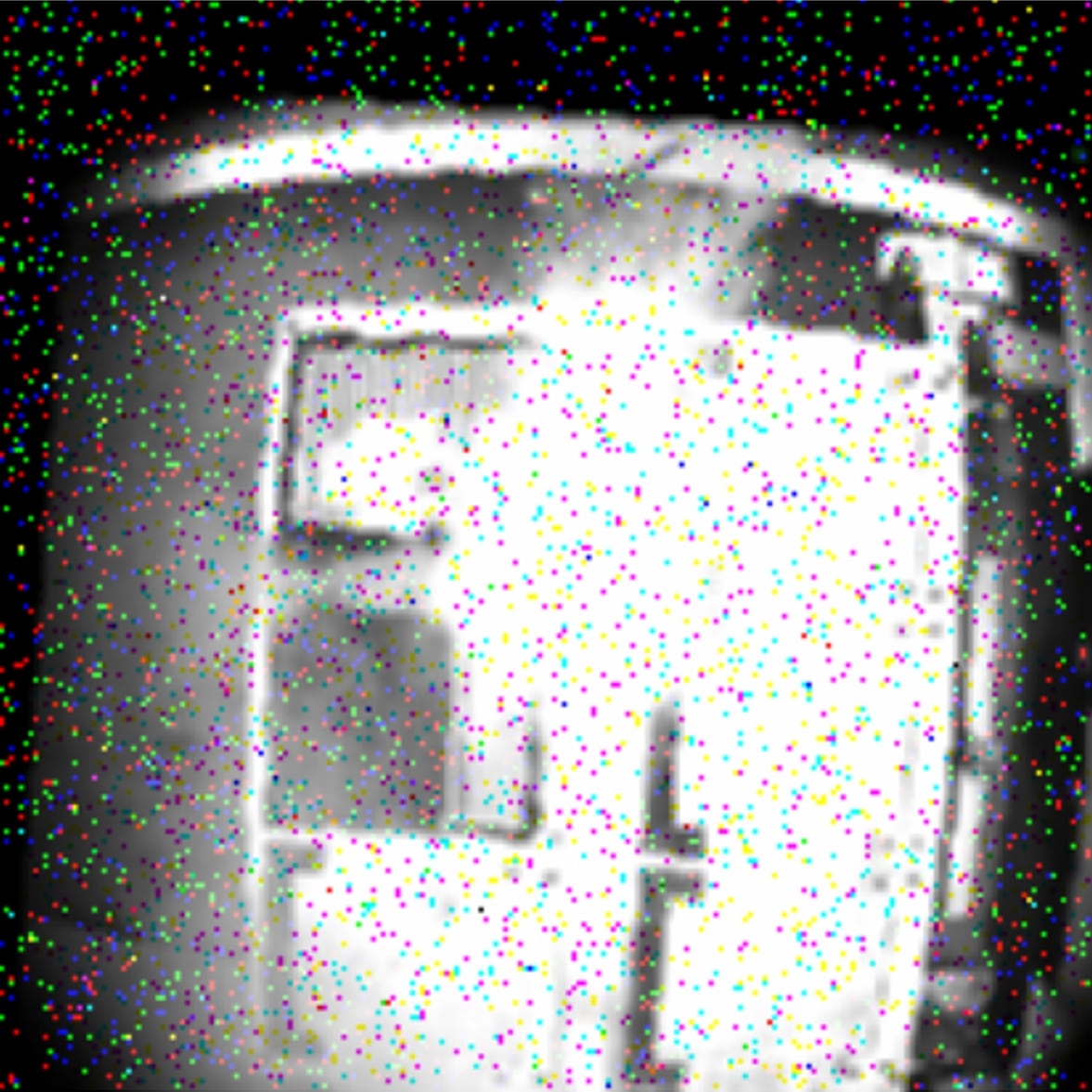}
    \hspace{0.2cm}
    \includegraphics[width=0.22\textwidth]{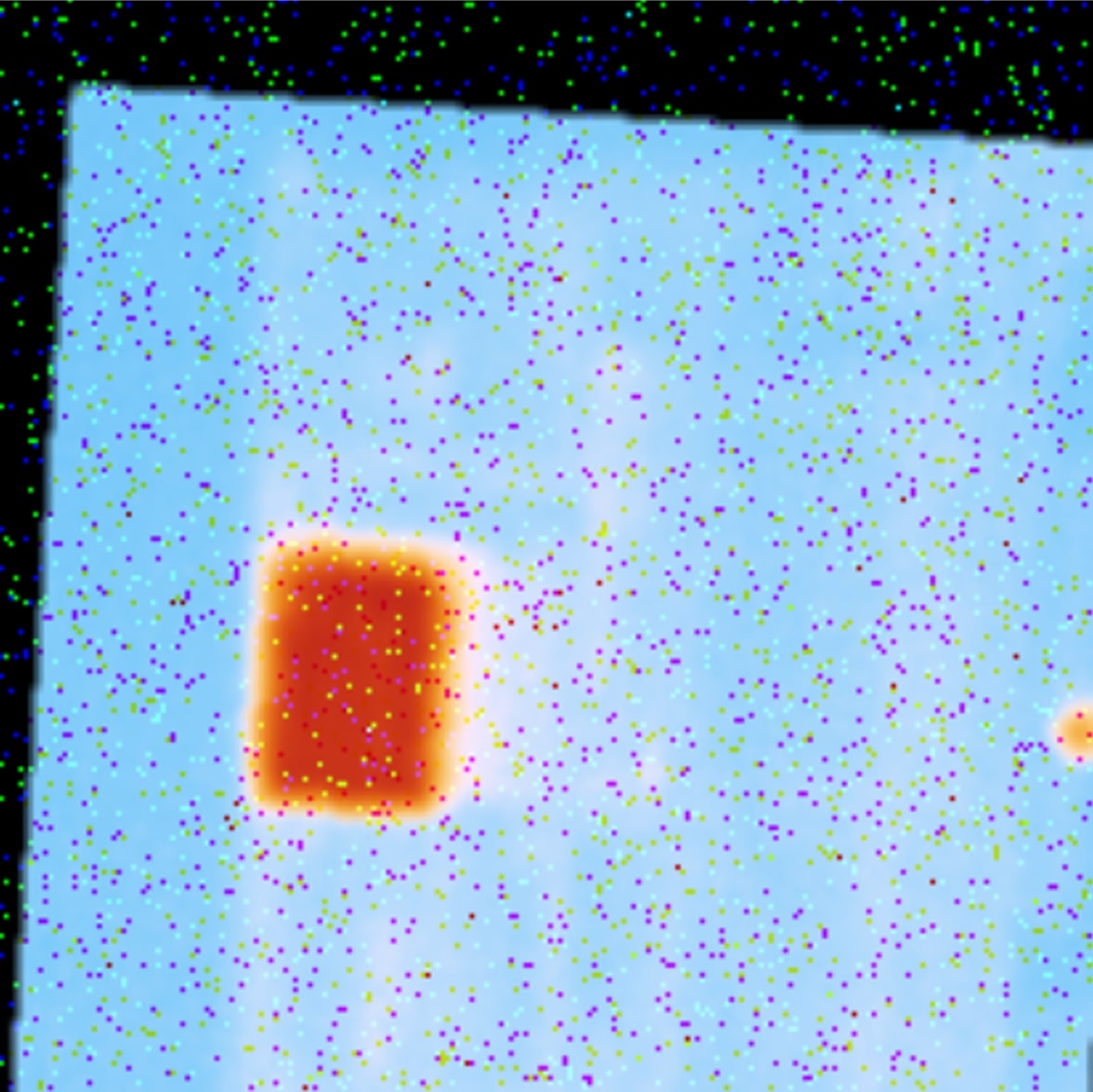}\\
    \vspace{0.1cm}
    (c)  \hspace{9.1cm} (d)
    \caption{Samples obtained using the same battery configuration. Optical and infrared are displayed in the left and right of each subfigure, respectively. Sample conditions are as follows: (a) baseline no augmentations; (b) thermal runaway no augmentations; (c) baseline heavy augmentations; (d) thermal runaway heavy augmentations. Note that (b) and (d) are distinct samples using different sized heat sources.}
    \label{fig:data}
\end{figure*}

In broad terms, the classification of images is a confluence of two components: the data and the model used to make the classification. In this section, we describe both of these components in sufficient detail to ensure reproducibility.
Furthermore, our data and implementation code has been made publicly available\footnote{\url{https://github.com/Athanath/Detect-Thermal-Runaway}} through Zenodo and GitHub, respectively.

\subsection{Data}

As described earlier, our thermal runaway detection models were developed in the context of a battery production line at VDL Nedcar. Specifically, it operates on a station early in the assembly process in which battery packs are positioned and mounted into their housing by a robotic arm, namely the ABB IRB 6700-245/3.00.
This arm is equipped with two downward-facing cameras overlooking the battery housing: the Cognex In-Sight 2800 optical camera and the DIAS PYROVIEW 128LS infrared camera. Custom software was developed for data collection which ensured that both cameras could be manually triggered. We note, however, that cameras were not triggered simultaneously and there was a delay of approximately 200 milliseconds between capturing optical and infrared images. As can be ascertained by the remainder of our data collection procedure, this delay is unlikely to have a significant impact on results.

Data collection was performed in two stages, corresponding to the baseline (non-thermal runaway) and thermal runaway cases. The former was relatively straightforward and merely involved capturing optical and infrared images for all possible configurations of the batteries in their housing. As the production line had two batteries and six possible positions in the housing, there were a total of 21 different configurations for when one or two batteries were housed. Twenty samples were obtained for each of the configurations, however, due to data corruption, our final dataset contained 412 baseline samples.
In order to collect thermal runaway data, the event had to be simulated. With safety considerations preventing real exothermic reactions, we simulated the features that would be captured by the two cameras during an actual event: heat and smoke. Heat was generated by using electric resistance heaters of different sizes as captured in Fig.~\ref{fig:data} (b) and (d). The devices were heated to a temperature greater than \SI{35}{\celsius} in our experiments. Smoke was simulated through the use of the ADJ VF1600 DMX fog machine. Specifically, smoke was injected at the source of the thermal runaway before image capture. In total 420 thermal runaway samples were obtained.
An example of a battery configuration under baseline and simulated thermal runaway conditions is showcased in Fig.~\ref{fig:data} (a) and (b), respectively. 

The data collection process yielded 832 total samples, a number considered small in the context of the models we use in our work \cite{d2020structural}. Perhaps the simplest way to overcome this is to generate a larger, upsampled dataset by drawing from the collected samples with replacement. The balanced class structure of our dataset allows for such a simple approach over more rigorous resampling techniques such as SMOTE \cite{chawla2002smote,dablain2022deepsmote} and ADASYN \cite{he2008adasyn} which focus on correcting a class imbalance. Mere upsampling is not sufficient to ensure the robustness of and prevent overfitting in our models, however, as it does not increase dataset diversity. Image augmentation is the process of applying transformations to existing images to artificially diversify the dataset and has been demonstrated as an effective way of improving model generalization \cite{taylor2018improving}. Many augmentation techniques have been proposed and several works have evaluated their downstream impact on deep learning models \cite{taylor2018improving,mikolajczyk2018data,xu2023comprehensive}.
Guided by these works, we developed a probabilistic augmentation pipeline to apply on both optical and infrared images. Specifically, the pipeline first included a reflection across the vertical axis followed by a 25 degree rotation. After this, a random affine transformation was applied, further rotating the image in addition to scaling and shearing. Gaussian blurring was then performed followed by salt-and-pepper noise.  
We note that as this was a probabilistic pipeline, each of its constituent steps had a $\sfrac{1}{2}$ chance of being performed with the exception of the Gaussian blur which was always performed.
The hyperparameters for each of these transformations can be found in our code. Examples of samples which have undergone augmentation are captured in Fig.~\ref{fig:data} (c) and (d).

\subsection{Models}

The optical and infrared images obtained for each sample in the data collection stage served as input to our models. One question that we were interested in was the potential for performance gain through the fusion of these two sources of information. As such, for each of the models investigated, we considered three cases: using only optical images, using only infrared images, and performing low-level data fusion on the two images to serve as input into our model. The low-level data fusion was performed such that the three channels of the optical image are concatenated with the three channels from the infrared image, resulting in a six channel representation of the original sample \cite{smolinska2019general}.
These input modalities lend themselves to deep learning models designed for processing and understanding image data. To this end, we chose to investigate three types of models: vanilla convolutional neural networks, residual neural networks, and vision transformers.
What follows is a brief description of each of these models.

\subsubsection{Convolutional Neural Network}

Convolutional neural networks (CNNs) \cite{cnn_original_paper} are the canonical deep learning approach to image classification. They rely on convolutional layers to extract spatial features by applying filters that detect patterns like edges, textures, or shapes in input data. These features are refined and combined through additional layers, often with non-linear activation functions followed by pooling layers which reduce dimensionality while retaining essential information. Finally, the features are fed into fully connected layers that combine them to make classifications.
Our work uses a shallow CNN with two convolutional layers and two fully-connected layers.
Although more advanced and better performing models have superseded such shallow CNNs \cite{szegedy2015going}, we include them in our work as their simple architecture allows for building and training the models ourselves, thus serving as a comparative baseline for the two pre-trained models we investigated and describe next.

\subsubsection{Residual Neural Network}

The residual neural network (ResNet) \cite{resnet_paper} is a type of CNN that was designed to address the vanishing gradient and degradation problems encountered when training deep neural networks \cite{denset_paper}. These problems are overcome by the introduction of residual connections that bypass one or more layers, allowing the network to learn the residual between the input and output of those layers. Several pre-trained variants of ResNet exist, bearing the name of the number of layers in their architecture. For instance, the ResnNet variant used in our work, ResNet-50 consists of 50 layers. The model is pre-trained on the ImageNet dataset \cite{deng2009imagenet} and thus only requires fine-tuning on our dataset by appending two fully connected layers to the pre-trained model.
When first introduced, ResNet achieved state-of-the-art results on image classification tasks and demonstrated the viability of training very deep neural networks. In recent years however, ResNet along with other pure CNN models have been outperformed by models which incorporate vision transformers into their architecture \cite{mauricio2023comparing,takahashi2024comparison}.

\subsubsection{Vision Transformers}

Vision transformers (ViTs) \cite{vit_original_paper} adapt the transformer architecture developed primarily in the field of natural language processing (NLP) for processing visual data. 
Unlike CNNs, ViTs divide an image into small patches, linearly embed them in a high-dimensional space, and feed the resulting sequence of embeddings into a transformer encoder akin to token embeddings in NLP. The model captures global context and relationships between all patches through multi-head self-attention mechanisms.
The ViT we evaluated in our work is the original architecture proposed by Dosovitskiy et al. \cite{vit_original_paper} pre-trained on the ImageNet-21k dataset \cite{ridnik2021imagenet21k} and fine-tuned on the aforementioned ImageNet dataset of Deng at al. As with ResNet, our ViT was further fine-tuned on our dataset. We note that the ViT implementation did not allow for six channelled images, thus we could not evaluate our fused dataset on this model.

\section{Evaluation}\label{sec:evaluation}

As alluded to earlier, the problem of thermal runaway detection is formulated as a classification task in which our models must predict a thermal runaway or non thermal runaway label given an input sample. As such, we use a standard procedure for evaluating evaluation wherein models are trained and tested on different subsets of data and compared through their performance on selected performance metrics. In our work these were the areas under the receiver operating characteristic (ROC) and precision-recall (PR) curves, denoted ROC-AUC and PR-AUC, respectively.
The ROC curve plots the true positive rate against the false positive rate by performing a sweep of potential threshold values, thus showing the trade-off between correctly identifying positives and mistakenly classifying negatives as positives. The ROC-AUC value ranges from zero to one, with one indicating perfect class separability and 0.5 indicating random guessing. Similarly, PR curves measure the trade-off between precision and recall across various thresholds. As with ROC-AUC, the values of PR-AUC range from zero to one with higher values indicating better performance.

\subsection{Evaluation Procedure}

Before training our models, the dataset was randomly partitioned into training, validation, and testing subsets constituting 70\%, 15\%, and 15\% of the samples, respectively. For each model, a grid search was performed through the corresponding set of model-specific hyperparameters. Specifically, for each hyperparameter combination in the search, the model was trained on the training subset and evaluated on the validation subset as per our performance metrics. The hyperparameter combination for each model deemed as best performing was then used to train the models on the training and validation subsets before predicting the testing subset. The performance metrics calculated on the testing subset were used for final model evaluation and comparison.

\subsection{Results}

\setlength{\tabcolsep}{0.4em}
\begin{table}[t!]
\caption{Model Performance on the Testing Subset as Per ROC-AUC and PR-AUC}
\begin{center}
\begin{tabular}{|c|c|c|c|c|c|}
\cline{2-6} 
\multicolumn{1}{c}{}&\multicolumn{3}{|c|}{\textbf{Dataset$^\text{a}$}} & \multicolumn{2}{|c|}{\textbf{Performance Metric}} \\
\hline
\multicolumn{1}{|c|}{\textbf{Model}} & Type & Upsampled &  Augmented  & ROC-AUC & PR-AUC \\
\hline
CNN & Optical & No & No &  1.00 & 1.00 \\
CNN & Infrared & No & No & 0.87 & 0.77 \\
CNN & Fusion & No & No & 1.00 & 1.00 \\
CNN & Optical & Yes & Yes & 0.72 & 0.38 \\
CNN & Infrared & Yes & Yes & 0.83 & 0.70 \\
CNN & Fusion & Yes & Yes & 1.00 & 1.00 \\
ResNet & Optical & Yes & Yes & 1.00 & 1.00 \\
ResNet & Infrared & Yes & Yes & 1.00 & 1.00 \\
ResNet & Fusion & Yes & Yes & 1.00 & 1.00 \\
ViT & Optical & Yes & Yes & 1.00 & 1.00 \\
ViT & Infrared & Yes & Yes & 1.00 & 1.00 \\
\hline 
\multicolumn{6}{l}{$^{\mathrm{a}}$Due to the small size of our dataset, only our CNN model was eval-} \\
\multicolumn{6}{l}{uated on the initial -- without upsampling or augmentation -- dataset.}
\end{tabular}
\label{tab:results}
\end{center}
\end{table}

\begin{figure*}[t!] 
    \centering
    \includegraphics[width=0.22\textwidth]{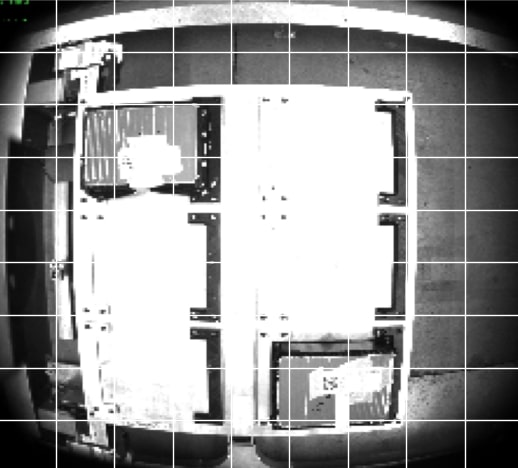}
    \hspace{0.2cm}
    \includegraphics[width=0.22\textwidth]{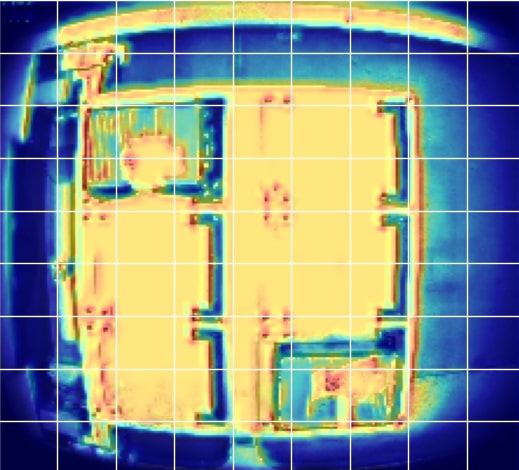}
    \hspace{1cm}
    \includegraphics[width=0.22\textwidth]{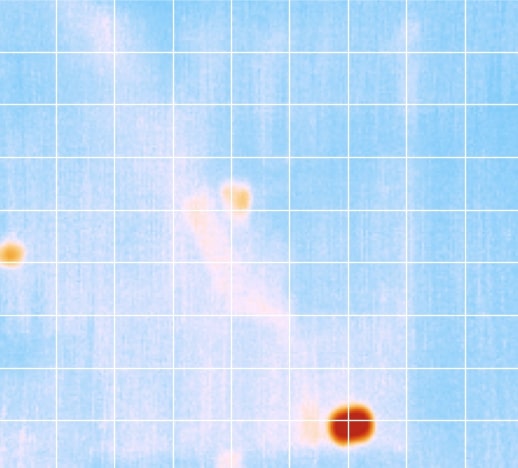}
    \hspace{0.2cm}
    \includegraphics[width=0.22\textwidth]{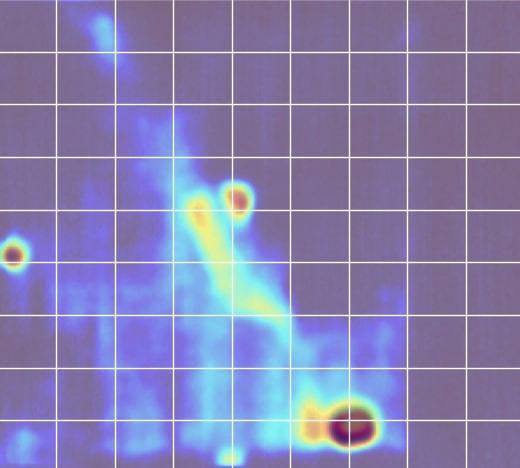}\\
    \vspace{0.1cm}
    (a) \hspace{9.1cm} (b) \\
    \vspace{0.2cm}
    \includegraphics[width=0.22\textwidth]{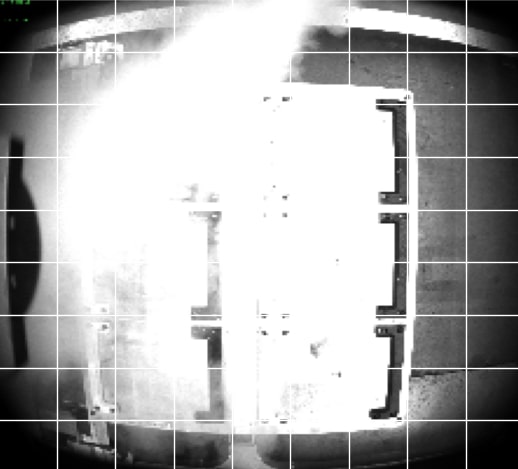}
    \hspace{0.2cm}
    \includegraphics[width=0.22\textwidth]{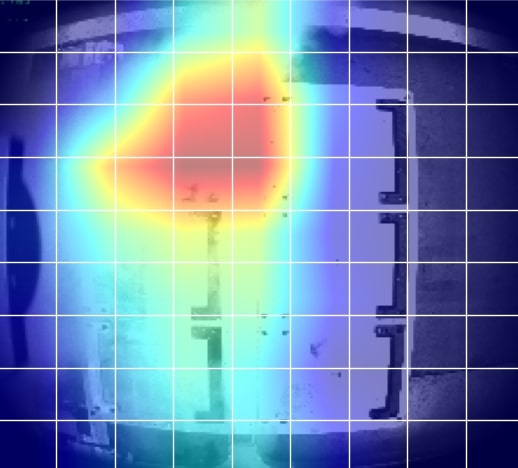}
    \hspace{1cm}
    \includegraphics[width=0.22\textwidth]{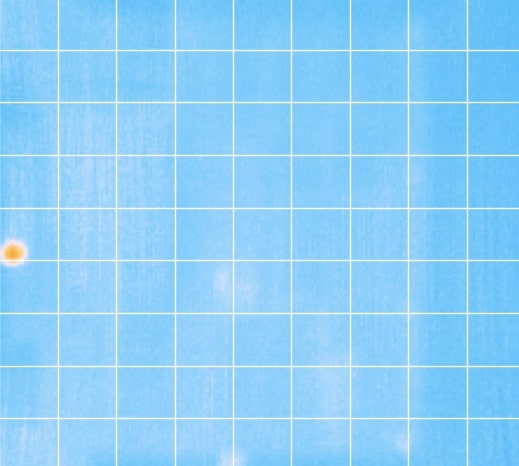}
    \hspace{0.2cm}
    \includegraphics[width=0.22\textwidth]{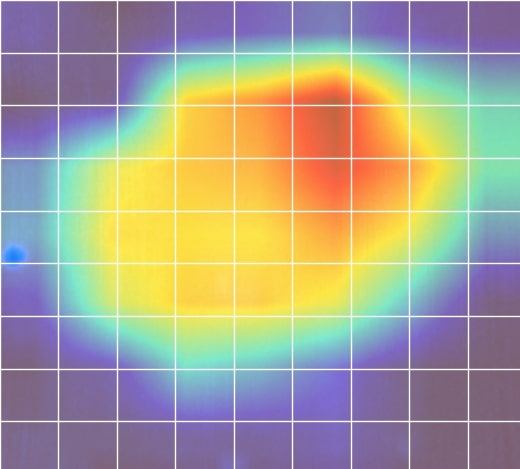}\\
    \vspace{0.1cm}
    (c)  \hspace{9.1cm} (d) \\
    \vspace{0.2cm}
    \includegraphics[width=0.22\textwidth]{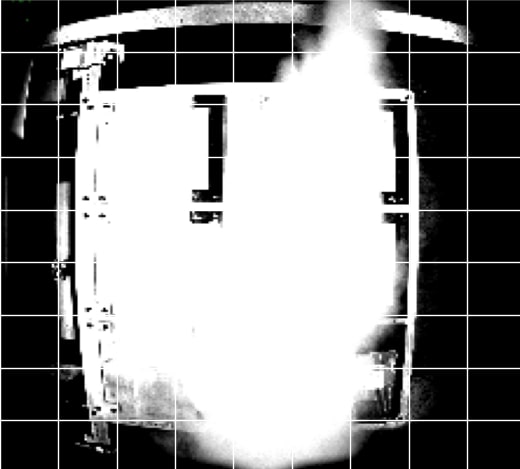}
    \hspace{0.2cm}
    \includegraphics[width=0.22\textwidth]{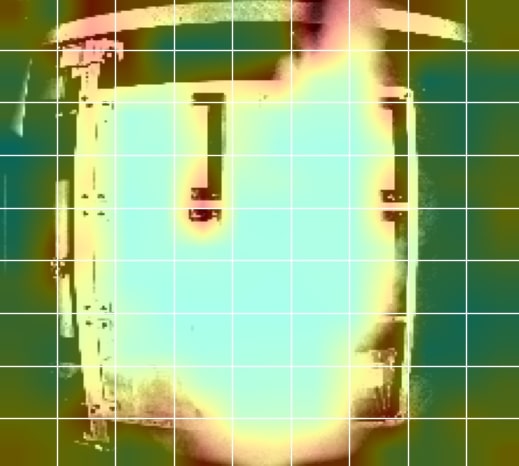}
    \hspace{1cm}
    \includegraphics[width=0.22\textwidth]{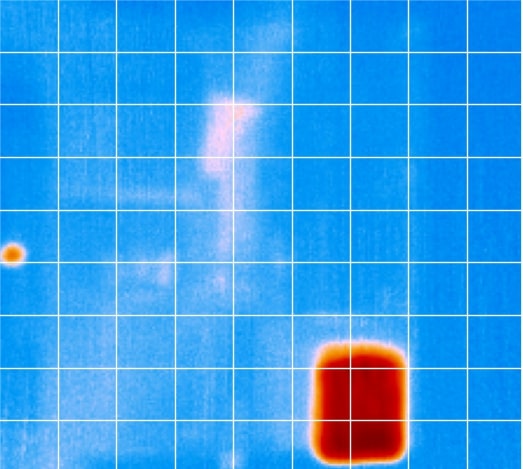}
    \hspace{0.2cm}
    \includegraphics[width=0.22\textwidth]{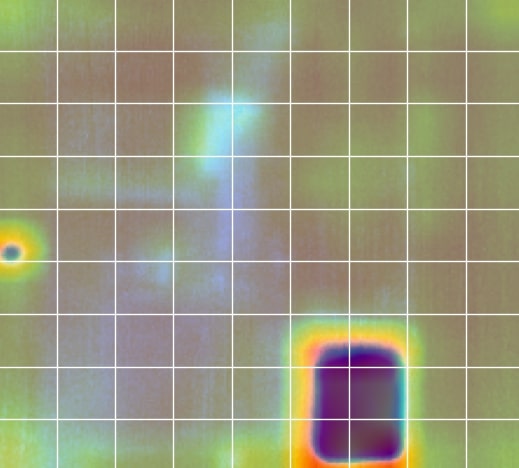}\\
    \vspace{0.1cm}
    (e)  \hspace{9.1cm} (f)
    \caption{
    Explanatory heatmaps obtained from our trained models on resampled and augmented images.
    Warmer colours indicate higher spatial importance for label prediction.
    Input image and heatmap are displayed on the left and right of each subfigure, respectively.
    Top, bottom, and middle rows correspond to explanations of CNN, ResNet, and ViT, respectively. Subfigures in the two leftmost columns display optical images whereas the two rightmost columns display infrared images. Explanations of fusion images may be found in our implementation code. 
    Sample labels are as follows: (a) baseline; (b) thermal runaway; (c) thermal runaway; (d) baseline; (e) thermal runaway; (f) thermal runaway.}
    \label{fig:explanations}
\end{figure*}

The ROC-AUC and PR-AUC results for our three models on the testing subset are summarized in Table \ref{tab:results}. We notice that all three models were able to achieve perfect ROC-AUC and PR-AUC performance, indicating strong class separability and demonstrating the viability of applying deep learning approaches for the purpose of thermal runaway detection in production lines. Such strong performance may indicate that our data collection process yielded a dataset too simple and finer distinctions between baseline and thermal runaway events could be captured by the models.
Nevertheless, we notice that our shallow CNN was not capable of perfect prediction in all dataset conditions. With this in mind, we conclude that the large, pre-trained models ResNet and ViT are better suited for thermal runaway detection.
Furthermore, we notice that data fusion may offer an advantage over solo optical and infrared inputs as highlighted in the performance gain by our CNN model on the upsampled and augmented dataset. Given the limited number of non-perfect results, a definitive conclusion about the utility of data fusion in this context cannot be made.
Finally, we note the runtime for performing prediction using our models. When performing CPU\footnote{Our experiments were run on an 11th Gen Intel(R) Core(TM) i7-1165G7 @ 2.80GHz, 2803 Mhz, 4 Core processor.} inference, the runtime is between 100 and 200 milliseconds for the CNN and between 300 and 500 milliseconds for the ResNet and ViT models.
Continuous learning -- the incremental retraining of the models as new data become available -- was not investigated in our work.


\subsection{Discussion}

Perhaps the least surprising outcome of our results is that deeper, pre-trained architectures outperformed shallower models trained from scratch. This is unsurprising as such models have more expressive architectures, enabling them to learn hierarchical and more complex features. Furthermore, they leverage the already learned robust feature representations from pre-training on large datasets. 
To aid in the interpretation of our models, we used two techniques for deep learning model explainability: gradient-weighted class activation mappings (Grad-CAM) and attention heatmaps.
Grad-CAM \cite{grad_cam_paper} is a visualization technique used to interpret CNNs by using the gradients of the target label flowing into the final convolutional layer to form weights which are then multiplied by the feature maps and summed to produce a heatmap. This heatmap indicates the most important regions in an image for label prediction. This result is visualized as an overlay on the input image, allowing for an easily interpretable explanation of the model's output. As this method was developed for CNNs, we were able to use it on our CNN and ResNet models. For ViT, we used the analogous attention heatmaps to explain the model. These heatmaps correspond to the attention weights that are calculated between each patch in the image and all other patches. This allows us to determine which areas of the image are most influential for predicting the labels by the model. The heatmaps obtained by our models on selected inputs are displayed in Fig.~\ref{fig:explanations}.
We noticed that all models were capable of identifying the relevant information -- smoke and heat for optical and infrared images, respectively -- to make classifications.  Noteworthy, however, is that correct classification was possible even when high importance was assigned to areas in an image that were not informative to the target label. For instance, ResNet exhibited a tendency to assign high importance to central areas in the image regardless of its contents as highlighted in Fig.~\ref{fig:explanations} (d).
Finally, a qualitative comparison of CNN explanations indicates that models trained on the upsampled and augmented datasets were better at identifying the importance regions in input images, despite the final classifications producing lower quality results as shown in Table \ref{tab:results}. 

The final issue we address is the relevance of the results to the context of a battery production line. Our models' perfect performance allows us to omit a discussion of the relative importance of false positives and false negatives in a production line context.
We note, however, that our dataset was collected to reflect a thermal runaway event at an advanced stage, once significant temperature rise and smoke buildup had occurred. Real-world systems would benefit from models trained and evaluated on data simulating thermal runaway at an earlier stage. Identifying thermal runaways earlier would allow for earlier triggering of safety procedures and thus give more time for proper handling of the event.
Finally, it should be noted that the models achieve a prediction runtime, which allows for application in the production line context and that our code is implemented for deployment compatibility with the ABB IRB communication protocols.

\section{Conclusion}\label{sec:conclusion}

In this paper, we investigated the use of deep learning methods for thermal runaway detection in battery production lines. To this end, we collected a dataset of baseline and thermal anomaly events simulated on a real-world production line. This dataset was then used to evaluate three common computer vision models. The results indicate that such models are a viable approach to detecting thermal runaway events in an industrial context. 
Despite these results, there are still avenues for future work. As already discussed, evaluating our models on data captured at an earlier stage of the thermal runaway event would allow for a better understanding of their sensitivity and ability to detect events at a time when more containment options are possible.
Furthermore, despite its compatibility, the integration of our models and the production line never came to fruition due to time and personnel constraints. Future work will thus involve the deployment of our models on the battery line. Finally, the prospect of continuous learning post-deployment presents an interesting road for further model development.

\section*{Acknowledgment}

This work has received financial support from the Ministry of Economic Affairs and Climate, under the grant ``R\&D Mobility Sectors'’ carried out by the Netherlands Enterprise Agency. We would like to thank VDL Nedcar and VDL Steelweld for their guidance, production line access, and providing this real-world use case.

\bibliographystyle{IEEEtran}
\bibliography{bibliography}

\end{document}